\documentclass[letterpaper, 10 pt, conference]{ieeeconf}  

\IEEEoverridecommandlockouts                              

\makeatletter
\let\NAT@parse\undefined
\makeatother

\usepackage{graphics} 
\usepackage{graphicx}
\usepackage{grffile} 
\usepackage{amsmath} 
\usepackage{amssymb}  
\usepackage{color}
\usepackage{bm}
\usepackage{multicol}
\usepackage{url}

\usepackage[noend]{algpseudocode}
\usepackage{algorithm}
\usepackage{makecell}
\usepackage{subfigure}
\usepackage[square, comma, numbers,sort&compress]{natbib}
\usepackage{multirow}
\let\labelindent\relax
\usepackage{enumitem}
\usepackage{caption}
\usepackage{pifont}
\usepackage{mwe}
\newcommand{\xmark}{\ding{55}}%
\usepackage[table,dvipsnames]{xcolor}
\usepackage[pagebackref=true,breaklinks=true,colorlinks,bookmarks=false]{hyperref}
\usepackage{nicematrix}
\usepackage{balance}

\hypersetup{
linkcolor=BrickRed
,citecolor=Green
,filecolor=Mulberry
,urlcolor=NavyBlue
,menucolor=BrickRed
,runcolor=Mulberry
,linkbordercolor=BrickRed
,citebordercolor=Green
,filebordercolor=Mulberry
,urlbordercolor=NavyBlue
,menubordercolor=BrickRed
,runbordercolor=Mulberry
}

\title{\LARGE \bf
NYC-Indoor-VPR: A Long-Term Indoor Visual Place Recognition Dataset with Semi-Automatic Annotation
}

\author{Diwei Sheng, Anbang Yang, John-Ross Rizzo, Chen Feng\textsuperscript{\ding{41}}\\
{\tt\small\url{https://ai4ce.github.io/NYC-Indoor-VPR}}
\thanks{New York University, Brooklyn, NY 11201, USA}%
\thanks{\ding{41} Corresponding author ({\tt\small \href{mailto:cfeng}{cfeng@nyu.edu}}). This work is supported by NSF Grant 2238968. We also thank the NYU HPC team for their assistance and support.}%
}

\IEEEaftertitletext{
\begin{center}
\vspace{-6mm}
            \captionsetup{type=figure}
            \includegraphics[width=0.9\textwidth]{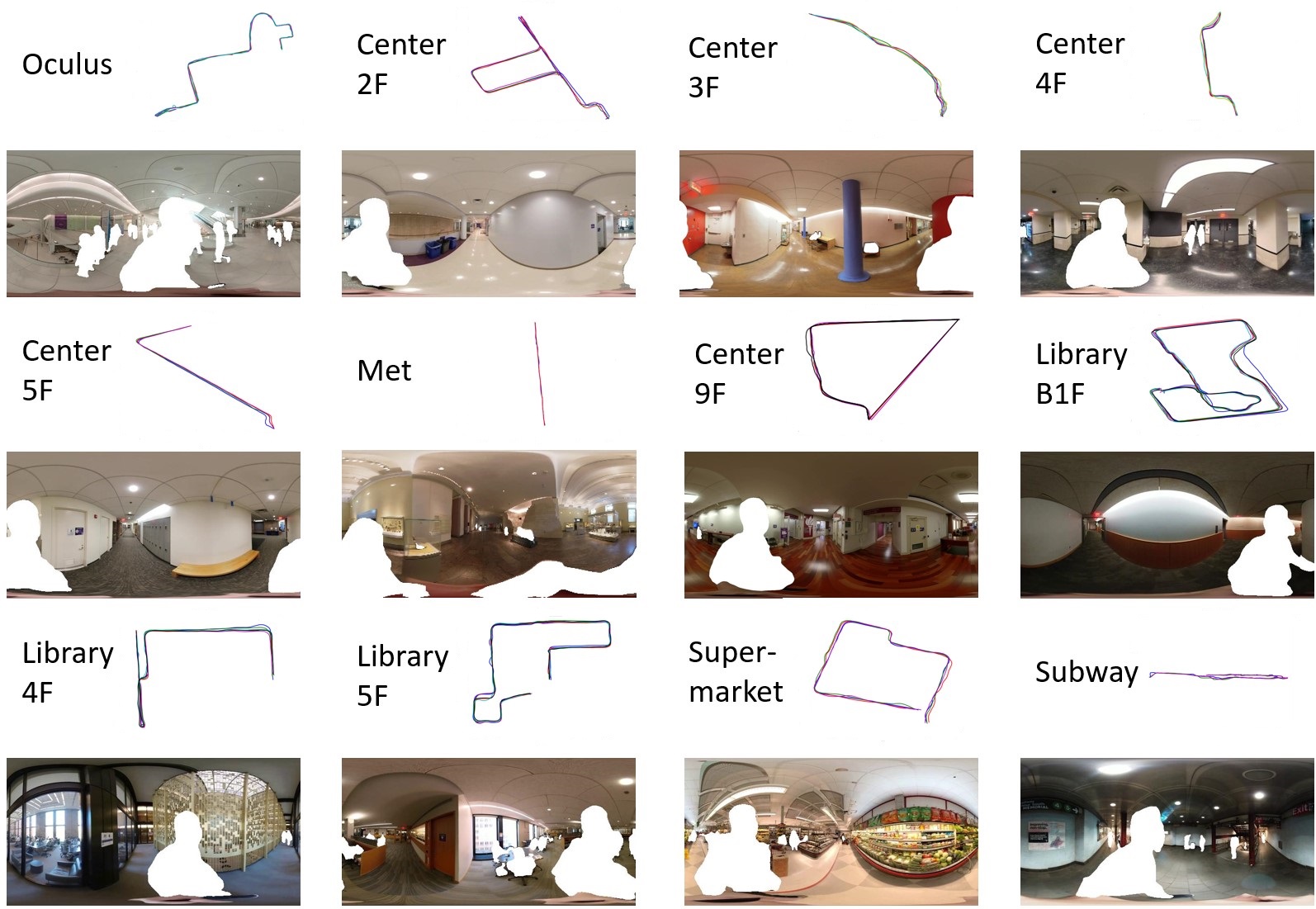}
            \captionof{figure}{
           Trajectories annotated by our semi-automatic method and example images of 12 scenes in NYC-Indoor-VPR.
           }
            \label{fig:data_vis}
\end{center}
}

\begin{document}

\maketitle
\thispagestyle{empty}
\pagestyle{empty}


\begin{abstract}

Visual Place Recognition (VPR) in indoor environments is beneficial to humans and robots for better localization and navigation. It is challenging due to appearance changes at various frequencies, and difficulties of obtaining ground truth metric trajectories for training and evaluation.
This paper introduces the NYC-Indoor-VPR dataset, a unique and rich collection of over $36,000$ images compiled from $13$ distinct crowded scenes in New York City taken under varying lighting conditions with appearance changes. \textit{Each scene has multiple revisits across a year}. To establish the ground truth for VPR, we propose a semiautomatic annotation approach that computes the positional information of each image. Our method specifically takes pairs of videos as input and yields matched pairs of images along with their estimated relative locations. The accuracy of this matching is refined by human annotators, who utilize our annotation software to correlate the selected keyframes. Finally, we present a benchmark evaluation of several state-of-the-art VPR algorithms using our annotated dataset, revealing its challenge and thus value for VPR research.

\end{abstract}

\section{Introduction}

Visual Place Recognition (VPR) enhances the ability of cyber-physical systems to recognize previously visited locations based on visual images. This is accomplished by comparing a given query image with a database of images, each associated with known camera positions. VPR applications extend across numerous sectors, including medical imaging, autonomous vehicles, assistive navigation for people with disabilities, and augmented reality. Both indoor and outdoor environments benefit from VPR, which provides accurate localization and navigation for users such as robots and vulnerable pedestrians with wearable computers. 

Indoor VPR, however, encounters unique challenges. Perceptual aliasing, where different places may appear visually identical, becomes an issue owing to structural repetition in buildings, such as hallways and rooms. Another difficulty arises from the obstruction of views in indoor environments, which are cluttered by walls, pillars, and moving objects. 

The progress of annotating the camera locations for the database images presents an additional obstacle for indoor VPR. These annotated locations function as ground truths, enabling the matching and localization of query images and the assessment of the VPR algorithm performance. In contrast to outdoor environments, where Global Positioning System (GPS) coordinates can be linked to database images, indoor environments often encompass several floors, rendering GPS coordinates inadequate for differentiating locations. Various datasets circumvent this limitation by employing laser scans or spherical cameras to produce a 3D point cloud of the environment~\cite{huitl2012tumindoor,sanchez2021risedb}. Others utilize sensors such as LiDARs and Inertial Measurement Units (IMUs) to track camera movements and locate images~\cite{sanchez2021risedb}.

\begin{figure}
    \centering
    \subfigure[COLMAP reconstruction]{
        \includegraphics[width=0.27\linewidth]{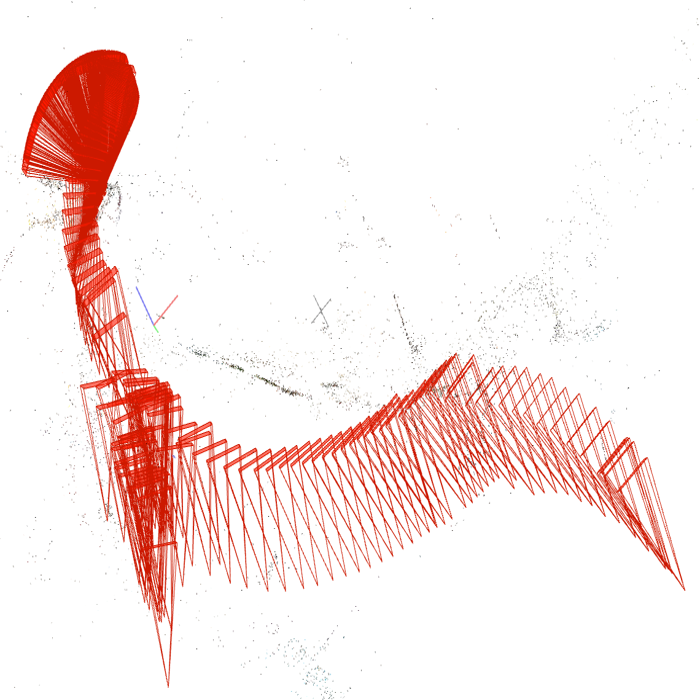}
        \label{fig:colmap_recons}
    }
    \subfigure[Visual SLAM topometric map]{
        \includegraphics[width=0.27\linewidth]{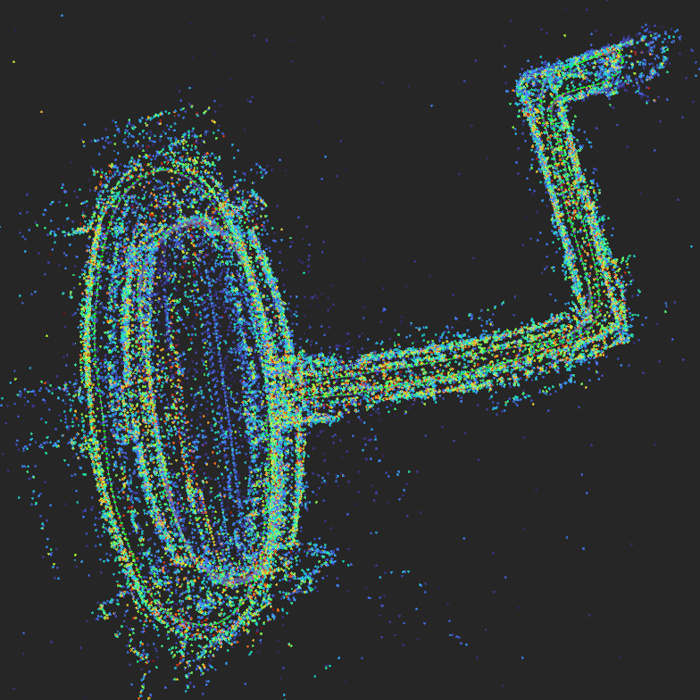}
        \label{fig:openvslam}
    }
    \subfigure[Semi-automatic annotation result]{
        \includegraphics[width=0.27\linewidth]{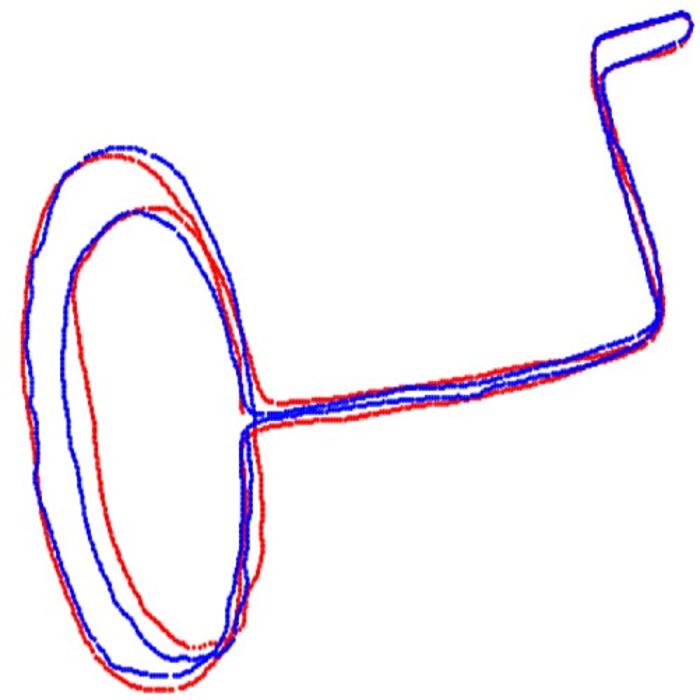}
        \label{fig:traj_pair}
    }
    \caption{Comparison of annotation methods for a video (pair) visiting Oculus. COLMAP fails to accurately reconstruct. Visual SLAM can generate a trajectory, but cannot match two trajectories. Our annotation method accurately computes the relative location of each frame in a video pair.}\label{fig:comp}
    \vspace{-3mm}
\end{figure}

Nevertheless, the inclusion of additional sensors increases the mapping costs. To alleviate this, several methods have been developed that use only image sequences to derive the relative locations of the images. Structure from Motion (SfM) techniques, such as COLMAP, involve extracting features from images, estimating camera poses and 3D points by matching these features across images and refining these estimations to reconstruct a 3D structure of the scene~\cite{schonberger2016structure}. However, SfM often fails to reconstruct large indoor scenes accurately (Fig.~\ref{fig:colmap_recons}) due to perceptual aliasing and blocked views. Furthermore, SfM reconstruction is redundant for VPR, which necessitates only the relative locations of the images and not a comprehensive 3D reconstruction of the environment. Simultaneous Localization and Mapping (SLAM) methods can generate a topometric map from an image sequence, as shown in Fig.~\ref{fig:openvslam}~\cite{murORB2}, but they fall short in matching two sequential trajectories, which are vital for VPR ground truth (Fig.~\ref{fig:traj_pair}). There is a clear gap in annotation methods that can calculate indoor image locations as the ground truth for VPR benchmarking effectively and accurately using only visual images.

In this paper, we propose a novel indoor VPR dataset and an associated benchmark. Our dataset includes a year-long collection of over $36,000$ images from $13$ different scenes captured using 360-degree cameras. Fig.~\ref{fig:data_vis} shows the trajectories and example images of certain scenes. We anonymize these images by whitening identity-related pixels, maintaining the privacy of the individual pedestrians. According to previous studies, the anonymization of pedestrians would not significantly affect the performance of existing VPR algorithms~\cite{sheng2021nyu}. Moreover, we propose a technique for generating ground-truth locations for our dataset, which enables us to examine the aforementioned challenges.

Our paper's key contributions are:
\begin{itemize}
\item The introduction of NYC-Indoor-VPR, \textit{a unique, year-long indoor VPR benchmark dataset} comprising images from different crowded scenes in New York City, taken under varying lighting conditions with appearance changes. This dataset, along with our benchmark code, is publicly available for research purposes. We also evaluate the performance of leading VPR algorithms.
\item The proposal of a semi-automatic annotation method that can efficiently and accurately match trajectories and generate images with topometric locations as ground truth, applicable to any indoor VPR dataset.
\end{itemize}

\section{Related Work}

\begin{table*}[t]
    \caption{Comparison of major public indoor VPR datasets with our NYC-Indoor-VPR.}
    \vspace{-3mm}
    \label{tab_1}
    \centering
    \resizebox{\textwidth}{!}{%
    \begin{tabular}{l|ccccccc}
    \hline
    Dataset     &   dynamic-object & crowded-area & anonymization & 360-view & \#images &\#scenes &time-span\\ \hline
    
    17 Places~\cite{sahdev2016indoor}& \xmark & \xmark      & \xmark & \xmark & 16,000 & 17 & 2 weeks\\\hline
    
    TUMindoor~\cite{huitl2012tumindoor}& \xmark & \xmark      & \xmark & \xmark & 41,888 & 7 & 1 month\\\hline
    
    RISEdb~\cite{sanchez2021risedb}& \xmark & \xmark      & \xmark & \checkmark & 1,000,000 & 5 & 3 months\\ \hline
    
    InLoc~\cite{taira2018inloc} & \checkmark & \checkmark & \xmark & \checkmark & 10,328 & 5 & months\\ \hline
    
    7 scenes~\cite{shotton2013scene} & \xmark & \xmark & \xmark & \xmark & 43,000 & 7 & unknown\\ \hline
     
    Baidu Mall~\cite{sun2017dataset} & \checkmark & \xmark & \xmark & \xmark & 682 & 1 & unknown\\ \hline
    
    Gardens Point~\cite{glover2014} & \xmark & \xmark & \xmark & \xmark & 400 & 1 & 1 day and 1 night\\ \hline
    
    \textbf{Ours} & \checkmark & \checkmark      & \checkmark & \checkmark & 36,107 & 13 & 1 year\\ \hline

    \end{tabular}%
    }
    \vspace{-5mm}
\end{table*}

This section covers related work in the areas of Indoor VPR datasets, annotation methods, and baseline methods.

\textbf{Indoor VPR datasets:} Given that NYC-Indoor-VPR is composed solely of indoor images for VPR, we compare it to other publicly available datasets with similar attributes. Key differences between these datasets and our proposed one are highlighted in Table~\ref{tab_1}. 

The presence of dynamic objects, such as pedestrians, complicates VPR owing to changes in the appearance and obstruction of the view. To maintain privacy in long-term datasets, pedestrians must be anonymized to prevent identity and potential spatiotemporal trajectory leaks. Thus, dataset collectors carefully control the presence of dynamic objects. For instance, RISEdb includes images from various buildings such as offices, conference venues, and restaurants~\cite{sanchez2021risedb}. Only a handful of datasets, like Baidu Mall and InLoc, depict people moving through a scene~\cite{sun2017dataset,taira2018inloc}. Crowded locations such as the World Trade Center have high pedestrian traffic. InLoc images were obtained from crowded university buildings. However, InLoc lacks anonymization. Our dataset employs MSeg, a semantic segmentation method, to isolate pedestrians and replace them with white pixels~\cite{lambert2020mseg}.

\begin{figure}
    \centering
    \subfigure[4 locations (row) with different appearances at different times.
    ]{
        \includegraphics[width=0.48\textwidth]{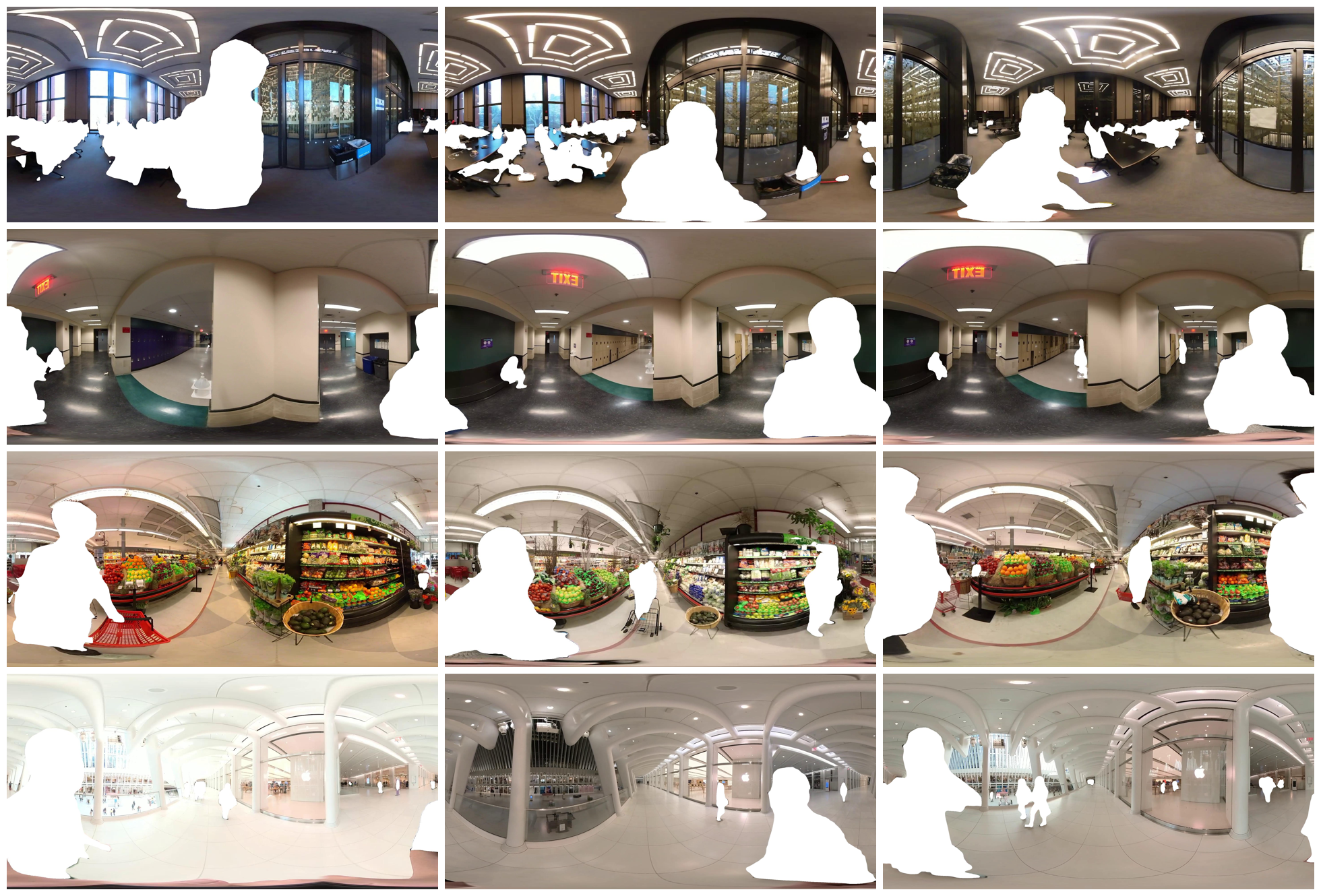}
        \label{fig:same_location}
    }
    \vspace{-3mm}
    \subfigure[Month distribution of visits]{
        \includegraphics[width=0.4\textwidth]{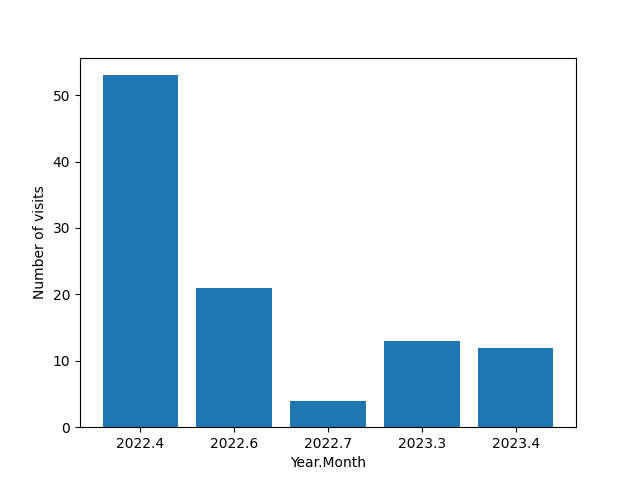}
        \label{fig:time_freq}
    }
    \caption{Our dataset is collected over a 1-year time span.}
    \vspace{-7mm}
\end{figure}

Matching images taken at the same location over an extended period is vital for VPR. Over time, image elements, such as illumination conditions, dynamic objects, and furniture distributions, have evolved. As seen in Table~\ref{tab_1}, Baidu Mall~\cite{sun2017dataset} and 7 scenes~\cite{shotton2013scene} do not explicitly mention the temporal difference between the database and query images. Other datasets, such as 17 Places~\cite{sahdev2016indoor}, TUMindoor~\cite{huitl2012tumindoor}, RISEdb~\cite{sanchez2021risedb}, and InLoc~\cite{taira2018inloc}, include images spanning weeks to months. NYC-Indoor-VPR covers a one-year timespan, offering a broader perspective on how varying appearances affect VPR, as shown by the time distribution of the images in Fig.~\ref{fig:time_freq}.

\textbf{Annotation methods:} To obtain ground truth, different methods are used to improve the efficiency and accuracy of human annotators. Baidu Mall used a three-step semi-automatic scheme to label the datasets~\cite{sun2017dataset}. InLoc applied a ‘sentinel’ task to safeguard annotation accuracy~\cite{taira2018inloc}. AnyLoc first gives annotators an existing dataset and subsequently shows differences to improve the accuracy~\cite{keetha2023anyloc}. In this paper, annotators utilize our custom annotation interface to match the identified keyframes. The trajectories are then matched, and frame pairs are extracted based on keyframe matching.

\textbf{Baseline methods:} Our selection includes state-of-the-art methods with varying architectures: ResNet+NetVLAD, Compact Convolutional Transformer (CCT)+NetVLAD, MixVPR, CosPlace, $R^{2}$ Former, and AnyLoc~\cite{arandjelovic2016netvlad, hassani2021escaping, ali2023mixvpr, berton2022rethinking, keetha2023anyloc, zhu2023r2former}. NetVLAD, a widely used aggregation method, inputs a dense feature map and outputs a vector of locally aggregated descriptors for the VPR. To derive dense feature maps from images, either a Convolutional Neural Network (CNN) backbone or a transformer backbone can be employed. We utilize ResNet-18 and CCT~\cite{he2016deep,hassani2021escaping} in our experiments. MixVPR views feature maps as a set of global features and establishes a global relationship among them~\cite{ali2023mixvpr}. CosPlace extracts distinctive descriptors from massive datasets~\cite{berton2022rethinking}. AnyLoc uses general-purpose feature representations derived from self-supervised models with no VPR-specific training~\cite{keetha2023anyloc}. Then it combines these derived features with unsupervised feature aggregation. $R^{2}$ Former is a unified place recognition framework that handles both retrieval and reranking with a novel transformer model. All of the chosen baseline methods are recent and have demonstrated competitive performance on large-scale datasets.

\section{Annotation method}
\begin{figure*}
    \centering
    \includegraphics[width=0.8\textwidth]{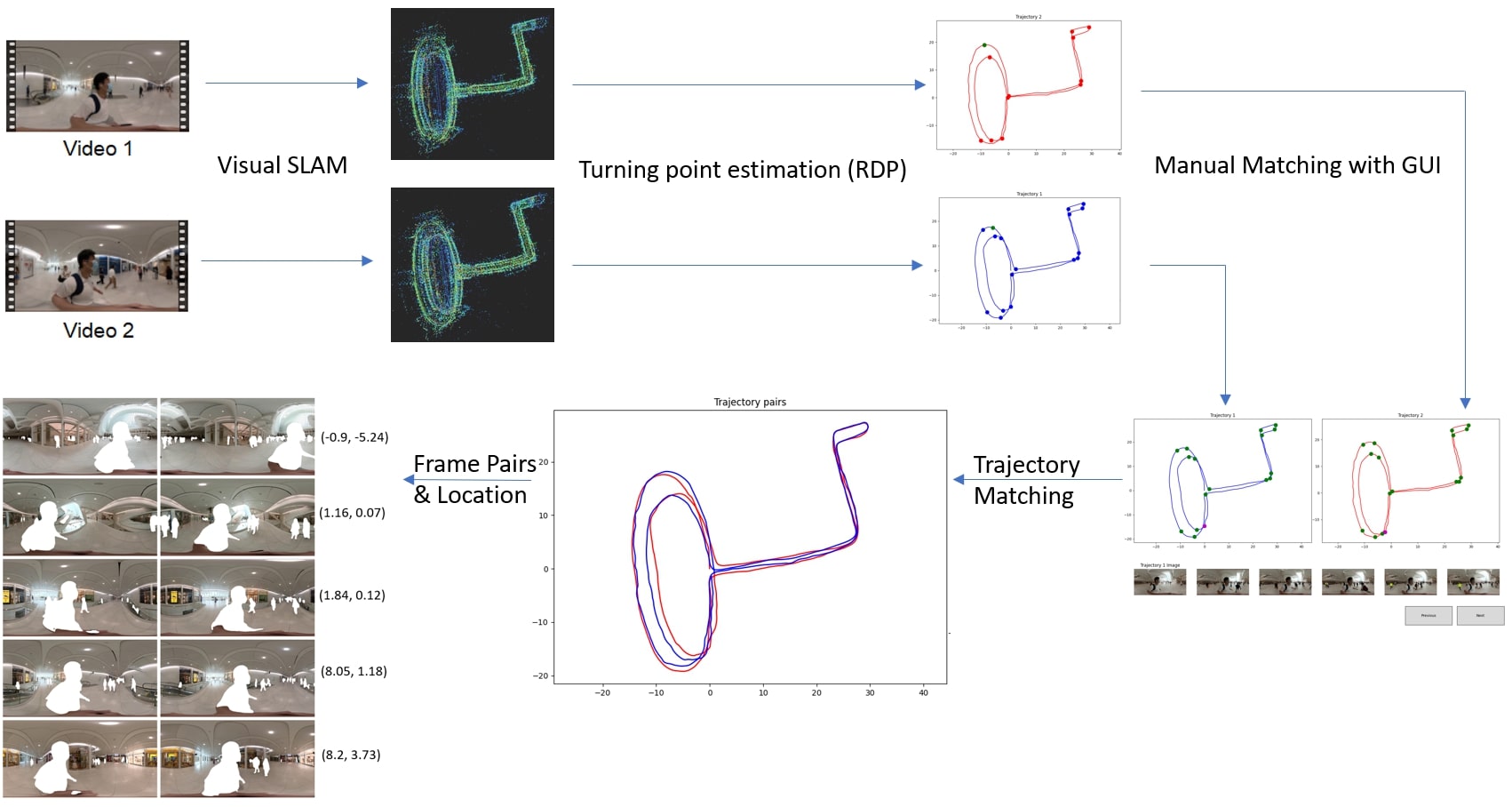}
    \caption{Overview of our semi-automatic annotation method. We collect two videos of the same route at different time. We use visual SLAM to identify keyframes with topometric locations. We automatically detect turning points (marked in green) and match them manually. We match the trajectory pairs and generate frame pairs with ground-truth topometric locations.}
    \label{fig:anno_arg}
\end{figure*}

Our aim is to generate images with topometric locations from indoor video trajectories. We view these topometric locations as the ground truth for the indoor dataset. Once established, we can use the dataset to benchmark the VPR methods. Given pairs of videos showing similar trajectories captured at different times, our method extracts frame pairs from the same location and computes their relative coordinates. By repeatedly pairing and annotating the videos with the same video, we created a dataset containing multiple images from the same location taken at different times.

Fig.~\ref{fig:anno_arg} presents the general pipeline of our annotation method. We delve deeper into each module below. 

\textbf{Visual SLAM:} Initially, we employ a variation of ORB-SLAM2 to estimate the topometric trajectory of the input video\footnote{\url{https://github.com/stella-cv/stella_vslam}}. ORB-SLAM2 is a visual SLAM framework known for its usability and extensibility~\cite{murORB2}. Using an equirectangular video, ORB-SLAM2 can accurately and robustly identify keyframes and their respective topometric locations.

\textbf{Segmentation and manual pairing:} Rather than directly matching all keyframes from trajectory pairs, we automatically detect and manually correct the trajectories' turning points for matching. This approach is adopted not only because turning points define the topometric shape of the trajectory but also because of the challenge of matching them owing to rapid rotation changes.

ORB-SLAM2 generates keyframe trajectories with topometric locations. We perform turning point estimation by first simplifying the trajectory using the Ramer-Douglas-Peucker algorithm (RDP). The RDP algorithm reduces the curve formed by the line segments into a similar curve with fewer points. It is noteworthy that points in the simplified curve do not necessarily originate from the original curve. We then iterate through the simplified piecewise curve to identify points with angles exceeding a set threshold. For each selected point, we find the nearest point in the original keyframe trajectory and designate it as the turning point.

We cannot simply match all turning points of a trajectory pair in sequence. Although we assume that the pair of trajectories starts and ends at the exact location and follows the same route, keyframes may not align perfectly because of differences in speed and trajectories, making their pairing uncertain. Also, One trajectory might turn earlier than the other at a crossroads or the estimated turning points may not coincide at a roundabout. We choose not to use floorplans to locate keyframes because floorplans are not available for every indoor environment. Without floorplans, human inspection of the generated ground truth is required to ensure accuracy under significant appearance changes. We involve human annotators who examine keyframes around the estimated matching turning point in the other trajectory for each turning point in one trajectory. The annotator then selects the keyframe with the most visually similar appearance, thus correcting the turning point match of the pair. In this way, we validate trajectory matching without sacrificing efficiency for additional inspection.


\textbf{Trajectory matching and ground truth generation:}
We match the trajectory pairs with the corrected turning point pairs. The transformation matrix is determined by solving the least-squares problem \(XA=Y\) where X and Y are turning point pairs. Upon trajectory matching, we extract frames from the video pair by assigning topometric locations as the ground truth. This is implemented by following the steps outlined below:

\begin{enumerate}
    \item Utilizing the timestamps of the turning points, we split the video pairs into pairs of segments.
    \item For each pair of segments with lengths \(T_1, T_2\), we determine the number of frames to be extracted, denoted by \(n = 2 \cdot \min(T_1, T_2)\).
    \item Both segments are then divided evenly into \(n\) parts to produce \(n\) frame pairs.
    \item To generate ground truth locations for the \(n\) frame pairs, we find the B-spline representation of all keyframe locations between turning points.
    \item We then evenly interpolate the B-spline curve to create \(n\) locations that correspond to the \(n\) frame pairs.
\end{enumerate}
In this way, we generate and add frames with ground-truth topometric locations to the dataset.

\section{The NYC-Indoor-VPR dataset}
\begin{table}[t]
    \caption{Dataset details.}
    \label{tab_2}
    \centering
    \resizebox{\linewidth}{!}{%
    \begin{tabular}{ll|cc}
    \hline
    Building     &   Scene(Floor) & \#visits & \#images\\ \hline
    
    The Oculus & floor 2 & 16 & 13933\\\hline
    Silver Center & floor 2 & 6 & 1580\\\hline
    Silver Center & floor 3 & 6 & 586\\\hline
    Silver Center & floor 4 & 6 & 940\\\hline
    Silver Center & floor 5 & 6 & 834\\\hline
    Silver Center & floor 6 & 6 & 814\\\hline
    Silver Center & floor 9 & 6 & 696\\\hline
    Bobst Library & floor -1 & 10 & 4044\\\hline
    Bobst Library & floor 4 & 10 & 3038\\\hline
    Bobst Library & floor 5 & 10 & 3847\\\hline
    Morton Williams Supermarket & floor 1 & 10 & 2237\\\hline
    Metropolitan Art Museum & floor 1 & 4 & 1266\\\hline
    Fulton Subway Station & floor 1 & 7 & 4627\\\hline

    \end{tabular}%
    }
\end{table}

The NYC-Indoor-VPR dataset comprises video frames recorded in New York City between April $2022$ and April $2023$. Footage was captured using hand-held Insta360 one x2 spherical cameras,  generating videos with a resolution of $1920\times 960$. We recorded images of $13$ different floors/scenes within the six buildings. Table~\ref{tab_2} presents details of the dataset. We chose buildings with varied utilities and appearances: the Oculus, New York University Silver Center for Arts and Science, Elmer Holmes Bobst Library, Morton Williams Supermarket, and Metropolitan Museum of Art. These settings represent a broad range of indoor spaces, including shopping malls, teaching buildings, libraries, supermarkets, and museums. 

For each building, we selected one or multiple floors as scenes. For each scene, we fixed the trajectory and captured videos along the same route at different times throughout the year. Fig.~\ref{fig:time_freq} shows the time distribution of visits. The videos were recorded from April to July $2022$ and from March to April $2023$. Therefore, it contains various changes in illumination and appearance. As shown in Fig.~\ref{fig:same_location}, we can see image changes at the same location over a year.

\begin{figure}
    \centering
    \includegraphics[width=\linewidth]{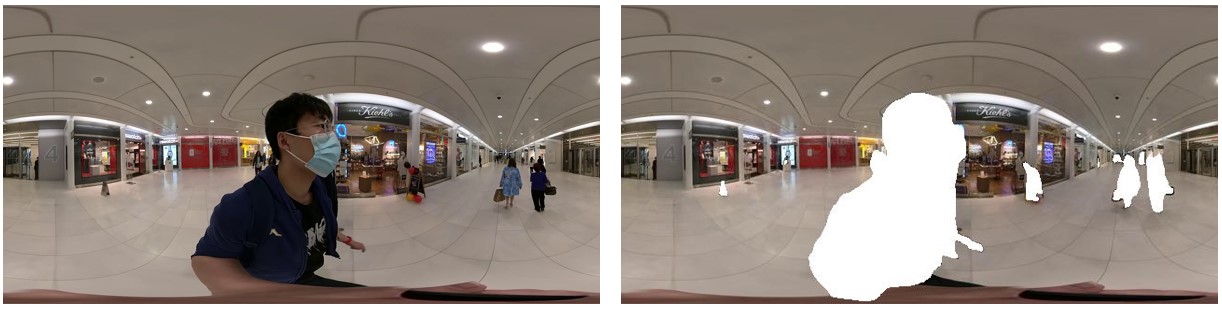}
    \caption{Raw image vs. Anonymized image}
    \label{fig:anony}
    \vspace{-3mm}
\end{figure}

We converted the 360-degree videos into an equirectangular format and then applied our semi-auto-annotation method to extract frame pairs from the video pairs in the same scene. Pedestrians were anonymized using MSeg~\cite{lambert2020mseg}, a semantic segmentation method that replaces them with white pixels. Fig.~\ref{fig:anony} shows the anonymized result of a dataset image.

\textbf{Uniqueness: } Our dataset stands out in two ways. First, NYC-Indoor-VPR images were captured in buildings such as The Oculus and the Bobst Library, which typically have a large flow of pedestrians. We anonymized these pedestrians in the images to reduce their exposure to personally identifiable information. These anonymized images not only enhance data privacy but also allow VPR algorithms to focus more on invariant or environmental features rather than transient features, such as moving people. Second, NYC-Indoor-VPR spans a year and includes images captured in buildings that undergo significant visual changes over time. For instance, goods in the supermarket vary and storefronts in the shopping mall are subject to change. This variability in the dataset allows us to test the performance of the VPR algorithms with fewer invariant features in the images.

\section{Benchmark Experiments}

\begin{figure*}[t]
    \centering
    \includegraphics[width=0.9\linewidth]{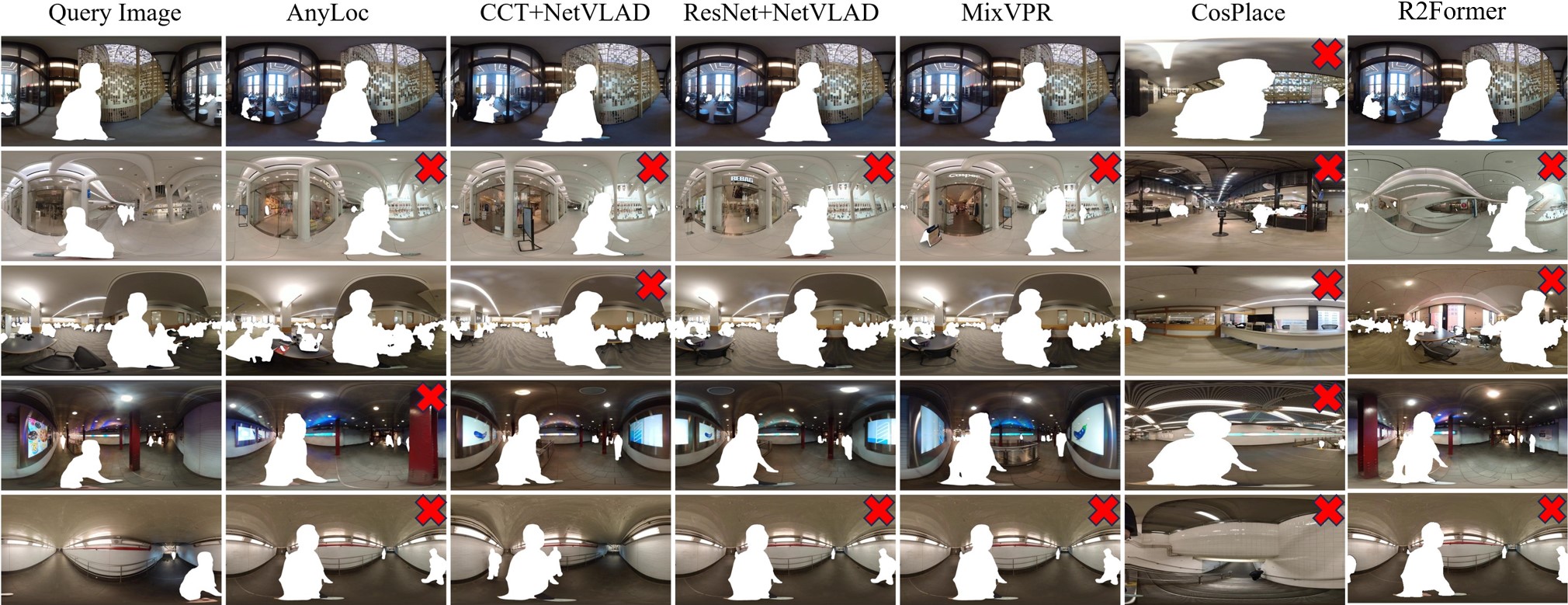}
    \caption{The VPR results for AnyLoc, CCT+NetVLAD, ResNet + NetVLAD, MixVPR, CosPlace, and $R^{2}$ Former are visualized. We randomly selected locations on Bobst Library's 4th floor, Oculus, Bobst Library's 5th floor, and Fulton subway station. The red cross indicates that the location of the retrieval image is not within the distance threshold (10 m). We show the top 1 retrieval for each method.}
    \label{fig:visual_result}
    \vspace{-3mm}
\end{figure*}

\subsection{Settings}
We benchmarked five state-of-the-art deep learning VPR methods on the NYC-Indoor-VPR dataset. We use nVidia RTX 2080S or Tesla V100 for all the experiments.

\textbf{Dataset: } For each scene, frame pairs in 2022 are used for training and validation and frame pairs in 2023 are used for testing. For each frame pair, one frame is considered as the database image, and the other is considered as the query.

\textbf{CosPlace: } CosPlace requires images with a certain field of view instead of panoramas as input. For each equirectangular image, we resize it to $1024\times 512$ and then cut it into four images, each with 90 °FoV. We train the model using a ResNet-18 with descriptors dimensionality of 512.

\textbf{MixVPR: } Because of the excellent transfer learning performance of the pre-trained model on datasets such as Pitts250k and Mapillary Street Level Sequences (MSLS), We directly use the model pre-trained on GSV-Cities dataset. The pretrained model has a ResNet50 backbone with a descriptor dimensionality of 4096.

\textbf{ResNet+NetVLAD: } We trained ResNet18 with NetVLAD end-to-end on our dataset. All the images are resized to $640\times 320$. The training and testing of NetVLAD require geometric coordinates. We replace the universal transverse mercator (UTM) with topometric coordinates generated by our semi-auto-annotation method.

\textbf{CCT+NetVLAD: } Compact Convolutional Transformer incorporates convolutional layers to insert the inductive bias of CNNs~\cite{hassani2021escaping}. We follow ~\cite{hassani2021escaping} and use the CLS token as a global descriptor, which is generated from the prepended learnable embedding of the sequence of patches. We resized all images to $384\times 384$ as required by CCT. The topometric coordinates are also used in this study.

\textbf{AnyLoc:} We choose the best performed AnyLoc-VLAD-DINOv2 from the paper. We follow the practices in the AnyLoc paper and use the ViT-G14 layer 31 value facet features, with 32 clusters for VLAD~\cite{keetha2023anyloc}. We resize all images to $640\times 320$. We directly use pretrained VLAD cluster centers. The global descriptor dimension generated for an image is 49152.

\textbf{Evaluation: } We follow ~\cite{berton2022deep} and use the metric of recall@N (R@N) to measure the percentage of queries for which one of the top-N retrieved images was taken within a certain distance of the query location. However, GPS coordinates are inaccurate in indoor environments due to signal obstruction. Instead of using GPS coordinates as in~\cite{berton2022deep}, we use topometric coordinates, which rank the retrieved database images by their relative distances to the query image. For each method, we measure R@N in each scene. Then we calculate the weighted average based on the number of images in each scene. In addition, we also experiment with splitting all the images directly into training, validation, and test sets without separating the scenes. The results are not significantly different from the sub-scene experiments.

\subsection{Results}

\begin{table}[ht]
    \caption{Retrieval results evaluated by Recall@N}
    \label{tab_3}
    \centering
    \resizebox{\linewidth}{!}{%
    \begin{tabular}{l|cccc}
    \hline
    Methods     &   R@1 & R@5 & R@10 & R@20\\ \hline
    
    CosPlace & 26.6 & 56.9 & 65.0 & 91.6\\\hline
    CCT+NetVLAD & 34.7 & 77.3 & 89.4 & \textbf{96.9}\\\hline
    ResNet+NetVLAD & \textbf{35.6} & 77.7 & 90.4 & 96.6\\\hline
    AnyLoc\footnotemark[2] & 27.6 & 69.8 & 84.9 & 95.6\\\hline
    MixVPR\footnotemark[2] & 34.4 & 76.9 & 89.6 & 96.0\\\hline
    $R^{2}$ Former\footnotemark[2] & 34.3 & \textbf{78.6} & \textbf{90.9} & 96.1\\\hline
    
    \end{tabular}%
    }
\end{table}

\footnotetext[2]{We directly use the pre-trained models of AnyLoc, MixVPR, and $R^{2}$ Former without training on our dataset because of their excellent transfer learning performance.}

\begin{figure}[t]
    \centering
    \includegraphics[width=\linewidth]{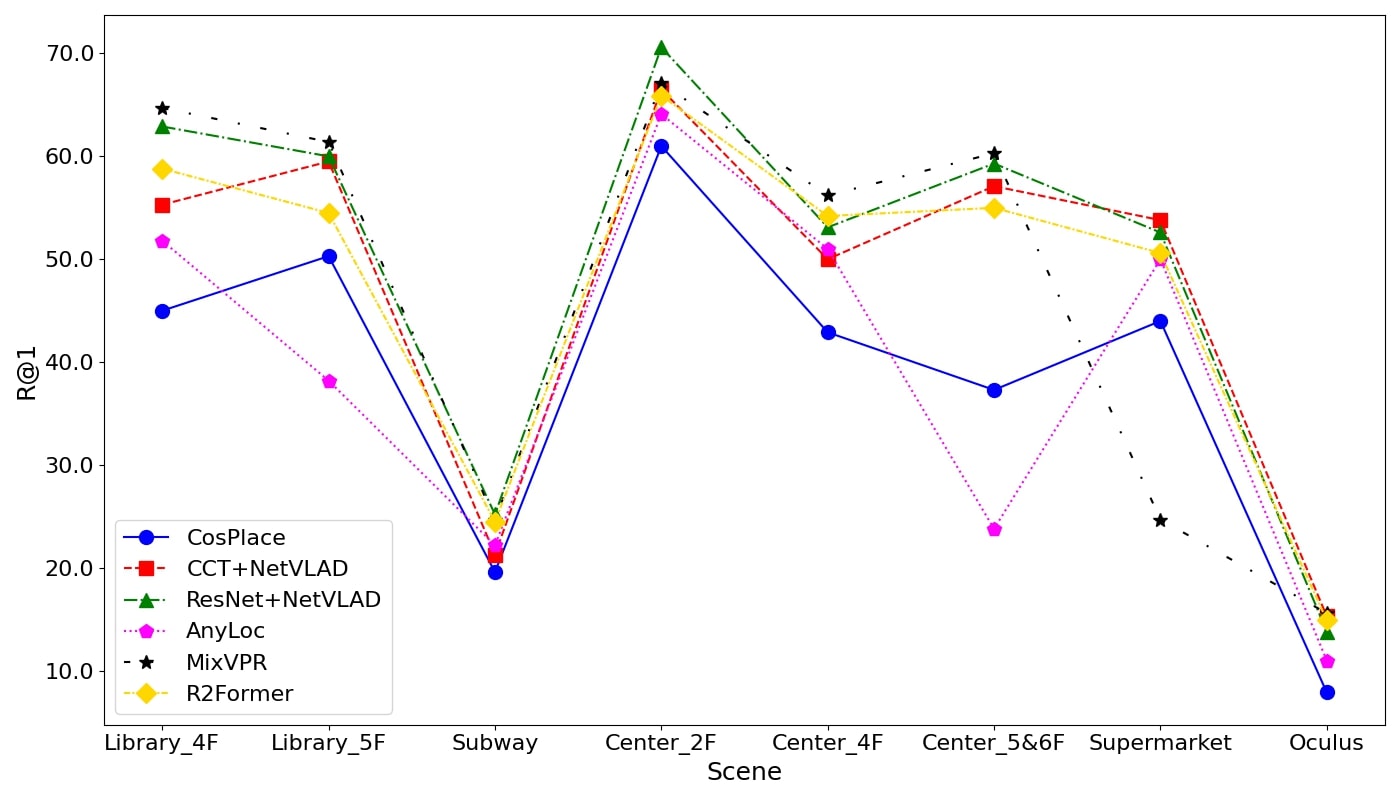}
    \vspace{-3mm}
    \caption{VPR success rate vs. query image scene.}
    \label{fig:scene}
    \vspace{-3mm}
\end{figure}

\textbf{Performance: }Table~\ref{tab_3} shows our main results for the performance of different VPR algorithms. $R^{2}$ Former has the highest recalls, followed by ResNet+NetVLAD, CCT+NetVLAD, MixVPR, and AnyLoc. The last method is CosPlace. One might be surprised by the relatively low performance of AnyLoc on NYC-Indoor-VPR since it achieved state-of-the-art performance on many large-scale indoor datasets such as Baidu Mall~\cite{sun2017dataset}, Gardens Point~\cite{glover2014}, and 17 Places~\cite{sahdev2016indoor}. AnyLoc outperforms NetVLAD, CosPlace, and MixVPR on these datasets~\cite{keetha2023anyloc}. However, compared to NYC-Indoor-VPR, these datasets lack characteristics such as crowded areas and equirectangular images, as shown in Table~\ref{tab_1}. We may attribute the low recall of AnyLoc to the unsuccessful representation learning of 360-view images and the view blocked by dynamic objects. Owing to the specific challenges for NYC-Indoor-VPR, AnyLoc is outperformed by NetVLAD, MixVPR, and $R^{2}$ Former. Cosplace is another method with near state-of-the-art performance on outdoor datasets, such as Tokyo247 and St Lucia~\cite{torii201524,milford2008mapping,berton2022rethinking}. Unlike MixVPR, the CosPlace model is trained on NYC-Indoor-VPR. However, CosPlace is designed for training extremely large datasets and casts training as a classification problem, rather than contrastive learning. Thus, Cosplace cannot capture subtle feature differences in the indoor environment. The visual results are shown in Fig.~\ref{fig:visual_result}. We can see that CosPlace performs worse than the other methods.

\textbf{Scene: } Fig.~\ref{fig:scene} shows the success rate of query images vs. different query image scenes. This result further confirms that CosPlace and AnyLoc perform worse than the other three methods in most scenes. The figure also clearly shows that the Fulton subway station and Oculus are challenging for the VPR. We hypothesize that the Fulton subway station contains hallways with repetitive features that cause perceptual aliasing, as shown in the bottom row of Fig.~\ref{fig:visual_result}. The Oculus also has hallways with pedestrian blocking features, as shown in the second row of Fig.~\ref{fig:visual_result}. The low VPR performance in these places demonstrates that NYC-Indoor-VPR contains the major challenges of indoor VPR, which are perceptual aliasing and obstruction of views.

\section{Conclusions}

In this paper, we propose a large-scale year-long indoor VPR dataset NYC-Indoor-VPR and an approach to facilitate the annotation of the relative coordinates of indoor image sequences. Our semi-auto-annotation method generates image pairs with ground-truth topometric locations from video pairs of the same trajectory. We demonstrate the necessity of our annotation methods compared to other methods such as SfM and SLAM. Our advantage is that it produces accurately matched trajectories with only a few keyframes matched by the human annotators. We applied our method to NYC-Indoor-VPR and used the annotated dataset to benchmark VPR algorithms. Experiments show that state-of-the-art VPR algorithms exhibit low performance owing to challenges in our dataset. Future work will include designing a VPR algorithm to address these challenges in indoor environments.



%




{\small
\bibliographystyle{IEEEtranN}
\balance
\bibliography{ai4ce-tpl}

\begin{thebibliography}{21}
\providecommand{\natexlab}[1]{#1}
\providecommand{\url}[1]{#1}
\csname url@samestyle\endcsname
\providecommand{\newblock}{\relax}
\providecommand{\bibinfo}[2]{#2}
\providecommand{\BIBentrySTDinterwordspacing}{\spaceskip=0pt\relax}
\providecommand{\BIBentryALTinterwordstretchfactor}{4}
\providecommand{\BIBentryALTinterwordspacing}{\spaceskip=\fontdimen2\font plus
\BIBentryALTinterwordstretchfactor\fontdimen3\font minus \fontdimen4\font\relax}
\providecommand{\BIBforeignlanguage}[2]{{%
\expandafter\ifx\csname l@#1\endcsname\relax
\typeout{** WARNING: IEEEtranN.bst: No hyphenation pattern has been}%
\typeout{** loaded for the language `#1'. Using the pattern for}%
\typeout{** the default language instead.}%
\else
\language=\csname l@#1\endcsname
\fi
#2}}
\providecommand{\BIBdecl}{\relax}
\BIBdecl

\bibitem[Huitl et~al.(2012)Huitl, Schroth, Hilsenbeck, Schweiger, and Steinbach]{huitl2012tumindoor}
R.~Huitl, G.~Schroth, S.~Hilsenbeck, F.~Schweiger, and E.~Steinbach, ``Tumindoor: An extensive image and point cloud dataset for visual indoor localization and mapping,'' in \emph{2012 19th IEEE International Conference on Image Processing}.\hskip 1em plus 0.5em minus 0.4em\relax IEEE, 2012, pp. 1773--1776.

\bibitem[Sanchez-Belenguer et~al.(2021)Sanchez-Belenguer, Wolfart, Casado-Coscolla, and Sequeira]{sanchez2021risedb}
C.~Sanchez-Belenguer, E.~Wolfart, A.~Casado-Coscolla, and V.~Sequeira, ``Risedb: a novel indoor localization dataset,'' in \emph{2020 25th International Conference on Pattern Recognition (ICPR)}.\hskip 1em plus 0.5em minus 0.4em\relax IEEE, 2021, pp. 9514--9521.

\bibitem[Schonberger and Frahm(2016)]{schonberger2016structure}
J.~L. Schonberger and J.-M. Frahm, ``Structure-from-motion revisited,'' in \emph{Proceedings of the IEEE conference on computer vision and pattern recognition}, 2016, pp. 4104--4113.

\bibitem[Mur-Artal and Tard\'os(2017)]{murORB2}
R.~Mur-Artal and J.~D. Tard\'os, ``{ORB-SLAM2}: an open-source {SLAM} system for monocular, stereo and {RGB-D} cameras,'' \emph{IEEE Transactions on Robotics}, vol.~33, no.~5, pp. 1255--1262, 2017.

\bibitem[Sheng et~al.(2021)Sheng, Chai, Li, Feng, Lin, Silva, and Rizzo]{sheng2021nyu}
D.~Sheng, Y.~Chai, X.~Li, C.~Feng, J.~Lin, C.~Silva, and J.-R. Rizzo, ``Nyu-vpr: long-term visual place recognition benchmark with view direction and data anonymization influences,'' in \emph{2021 IEEE/RSJ International Conference on Intelligent Robots and Systems (IROS)}.\hskip 1em plus 0.5em minus 0.4em\relax IEEE, 2021, pp. 9773--9779.

\bibitem[Sahdev and Tsotsos(2016)]{sahdev2016indoor}
R.~Sahdev and J.~K. Tsotsos, ``Indoor place recognition system for localization of mobile robots,'' in \emph{2016 13th Conference on computer and robot vision (CRV)}.\hskip 1em plus 0.5em minus 0.4em\relax IEEE, 2016, pp. 53--60.

\bibitem[Taira et~al.(2018)Taira, Okutomi, Sattler, Cimpoi, Pollefeys, Sivic, Pajdla, and Torii]{taira2018inloc}
H.~Taira, M.~Okutomi, T.~Sattler, M.~Cimpoi, M.~Pollefeys, J.~Sivic, T.~Pajdla, and A.~Torii, ``Inloc: Indoor visual localization with dense matching and view synthesis,'' in \emph{Proceedings of the IEEE Conference on Computer Vision and Pattern Recognition}, 2018, pp. 7199--7209.

\bibitem[Shotton et~al.(2013)Shotton, Glocker, Zach, Izadi, Criminisi, and Fitzgibbon]{shotton2013scene}
J.~Shotton, B.~Glocker, C.~Zach, S.~Izadi, A.~Criminisi, and A.~Fitzgibbon, ``Scene coordinate regression forests for camera relocalization in rgb-d images,'' in \emph{Proceedings of the IEEE conference on computer vision and pattern recognition}, 2013, pp. 2930--2937.

\bibitem[Sun et~al.(2017)Sun, Xie, Luo, and Wang]{sun2017dataset}
X.~Sun, Y.~Xie, P.~Luo, and L.~Wang, ``A dataset for benchmarking image-based localization,'' in \emph{Proceedings of the IEEE Conference on Computer Vision and Pattern Recognition}, 2017, pp. 7436--7444.

\bibitem[Glover(2014)]{glover2014}
A.~Glover, ``Gardens point day and night, left and right,'' \emph{Zenodo DOI}, vol.~10, 2014.

\bibitem[Lambert et~al.(2020)Lambert, Liu, Sener, Hays, and Koltun]{lambert2020mseg}
J.~Lambert, Z.~Liu, O.~Sener, J.~Hays, and V.~Koltun, ``Mseg: A composite dataset for multi-domain semantic segmentation,'' in \emph{Proceedings of the IEEE/CVF conference on computer vision and pattern recognition}, 2020, pp. 2879--2888.

\bibitem[Keetha et~al.(2023)Keetha, Mishra, Karhade, Jatavallabhula, Scherer, Krishna, and Garg]{keetha2023anyloc}
N.~Keetha, A.~Mishra, J.~Karhade, K.~M. Jatavallabhula, S.~Scherer, M.~Krishna, and S.~Garg, ``Anyloc: Towards universal visual place recognition,'' \emph{arXiv preprint arXiv:2308.00688}, 2023.

\bibitem[Arandjelovic et~al.(2016)Arandjelovic, Gronat, Torii, Pajdla, and Sivic]{arandjelovic2016netvlad}
R.~Arandjelovic, P.~Gronat, A.~Torii, T.~Pajdla, and J.~Sivic, ``Netvlad: Cnn architecture for weakly supervised place recognition,'' in \emph{Proceedings of the IEEE conference on computer vision and pattern recognition}, 2016, pp. 5297--5307.

\bibitem[Hassani et~al.(2021)Hassani, Walton, Shah, Abuduweili, Li, and Shi]{hassani2021escaping}
A.~Hassani, S.~Walton, N.~Shah, A.~Abuduweili, J.~Li, and H.~Shi, ``Escaping the big data paradigm with compact transformers,'' \emph{arXiv preprint arXiv:2104.05704}, 2021.

\bibitem[Ali-Bey et~al.(2023)Ali-Bey, Chaib-Draa, and Giguere]{ali2023mixvpr}
A.~Ali-Bey, B.~Chaib-Draa, and P.~Giguere, ``Mixvpr: Feature mixing for visual place recognition,'' in \emph{Proceedings of the IEEE/CVF Winter Conference on Applications of Computer Vision}, 2023, pp. 2998--3007.

\bibitem[Berton et~al.(2022{\natexlab{a}})Berton, Masone, and Caputo]{berton2022rethinking}
G.~Berton, C.~Masone, and B.~Caputo, ``Rethinking visual geo-localization for large-scale applications,'' in \emph{Proceedings of the IEEE/CVF Conference on Computer Vision and Pattern Recognition}, 2022, pp. 4878--4888.

\bibitem[Zhu et~al.(2023)Zhu, Yang, Chen, Shah, Shen, and Wang]{zhu2023r2former}
S.~Zhu, L.~Yang, C.~Chen, M.~Shah, X.~Shen, and H.~Wang, ``R2former: Unified retrieval and reranking transformer for place recognition,'' in \emph{Proceedings of the IEEE/CVF Conference on Computer Vision and Pattern Recognition}, 2023, pp. 19\,370--19\,380.

\bibitem[He et~al.(2016)He, Zhang, Ren, and Sun]{he2016deep}
K.~He, X.~Zhang, S.~Ren, and J.~Sun, ``Deep residual learning for image recognition,'' in \emph{Proceedings of the IEEE conference on computer vision and pattern recognition}, 2016, pp. 770--778.

\bibitem[Berton et~al.(2022{\natexlab{b}})Berton, Mereu, Trivigno, Masone, Csurka, Sattler, and Caputo]{berton2022deep}
G.~Berton, R.~Mereu, G.~Trivigno, C.~Masone, G.~Csurka, T.~Sattler, and B.~Caputo, ``Deep visual geo-localization benchmark,'' in \emph{Proceedings of the IEEE/CVF Conference on Computer Vision and Pattern Recognition}, 2022, pp. 5396--5407.

\bibitem[Torii et~al.(2015)Torii, Arandjelovic, Sivic, Okutomi, and Pajdla]{torii201524}
A.~Torii, R.~Arandjelovic, J.~Sivic, M.~Okutomi, and T.~Pajdla, ``24/7 place recognition by view synthesis,'' in \emph{Proc. IEEE Conf. Computer Vision and Pattern Recognition (CVPR)}, 2015, pp. 1808--1817.

\bibitem[Milford and Wyeth(2008)]{milford2008mapping}
M.~J. Milford and G.~F. Wyeth, ``Mapping a suburb with a single camera using a biologically inspired slam system,'' \emph{IEEE Transactions on Robotics}, vol.~24, no.~5, pp. 1038--1053, 2008.

\end{thebibliography}
}

\end{document}